# NAQD Env
# A benchmark for selective withdrawal in language agents

Mohamed Abouzahra
California State University, Monterey Bay

September 28, 2026

## Abstract

Language agents must revise planned actions when evidence changes, permission is revoked, or a stop instruction arrives. A useful response is selective: suspend affected actions, preserve unaffected work, and resume only after sufficient repair. We introduce NAQD-Env, a synthetic environment that evaluates these decisions against a deterministic reference policy over explicit evidence, authorization, and constraint dependencies. Eleven dependency families support evaluation on development structures, held-out families, and held-out combinations of structures. Metrics distinguish attempted violations from violations permitted by a simulated execution gate and jointly report policy agreement, task value, withdrawal, resumption, and event reporting. We evaluate three open-weight instruction-tuned models from two families under three prompt conditions on 350 frozen scenarios, yielding 3,150 model–prompt episodes before gate replay. Across the reported conditions, withdrawal recall is at most 0.06, no valid resumption is observed at eligible opportunities, and only one episode matches the complete reference policy. Under the NAQD prompt, Qwen2.5-7B has fewer unsafe-attempt episodes than Qwen2.5-3B and Llama-3.1-8B, but also completes less useful work and preserves unaffected actions less accurately. Exploratory supervised fine-tuning probes increase Qwen2.5-3B decision accuracy from 0.45–0.54 to 0.83–0.92; separate diagnostics reveal inappropriate withdrawal after curriculum omissions and a loss of event reporting. These results motivate evaluating selective withdrawal as a distinct component of agent reliability. The setting measures policy application with trusted structured inputs and does not establish real-world containment or source-verification ability.

## 1 Introduction

An agent that books travel, files purchase orders, or shares clinical documents acts on premises: a supplier's certification is current, a budget is approved, or a grant permits sending a document to a partner. These premises can change after a plan is prepared. A check may return a negative result, an issuer may revoke a grant, or a supervisor may issue a stop instruction. Selective withdrawal requires the agent to suspend the actions that have lost support, preserve unaffected work, and resume only after the relevant blocking causes have been repaired. Truth-maintenance systems and belief-revision theory provide foundations for maintaining dependencies under change [Doyle, 1979; de Kleer, 1986; Alchourrón et al., 1985]. For language agents, related research examines prompt injection, interruption, memory revision, and failure attribution. Our focus is the action-level suspension and resumption decision under an explicit policy.

Three requirements shape this evaluation. First, scoring requires a specification of which actions depend on which premises and how events change those premises. Second, the environment must represent structures that defeat simple mention matching: surviving alternative evidence, recursive delegation, and combinations of individually permitted actions that are jointly prohibited. Third, reporting safety alongside useful task completion is necessary because withholding all actions can reduce violations without correctly maintaining the plan. Separating proposed actions from simulated execution also prevents a controller's protection from being attributed to the model.

We present NAQD-Env, an environment built around these three requirements, and use it to establish baselines for three open models from two families under three prompt conditions on three frozen regimes. Our contributions:

A typed-premise environment with deterministic reference decisions. Actions depend on evidential and authorizational conditions and the absence of impediments. Eleven dependency families cover direct support, propagation, alternatives, delegation, and prohibited combinations. The reference policy maps each action and decision point to a prescribed option.

A multidimensional scoring protocol. Attempted violations, simulated executed violations, policy agreement, task value, and event reporting are reported separately. Withdrawal precision and recall distinguish missed suspensions from unnecessary suspensions, and resumption is conditioned on an eligible prior withdrawal. Protocol defects remain visible alongside retained valid decisions.

A versioned evaluation instrument. Scripted controls, regression tests, checksummed scenario builds, enforced split boundaries, and recorded replies support verification and reanalysis. These checks establish specific implementation properties; they do not prove freedom from every form of leakage or validate the policy for every deployment.

Baselines and exploratory training probes. Three models from two families are evaluated under three prompt conditions on 350 frozen cases each. The results distinguish persistent execution from verification and refusal, and illustrate how aggregate accuracy can conceal failures in withdrawal, repair, and reporting.

The contribution is an evaluation task and instrument for structured policy application. Its reference semantics are explicit design choices, and its conclusions are limited to the tested models, scenarios, and inference configuration.

## 2 Related work

Agent safety benchmarks. AgentDojo [Debenedetti et al., 2024] and InjecAgent [Zhan et al., 2024] evaluate indirect prompt injection in tool-using agents. ToolEmu [Ruan et al., 2023] uses language models to emulate tool execution and assess risks, while AgentHarm [Andriushchenko et al., 2024] evaluates harmful task completion as well as refusal. These benchmarks already connect safety with task behavior. NAQD-Env adds a controlled suspension and repair task over explicitly enumerated action dependencies; it does not claim that joint safety and utility evaluation is new.

Shutdown and intervention. Schlatter et al. [2025] study models that interfere with shutdown during incomplete tasks. StepShield [Felicia et al., 2026] evaluates the timing of monitor intervention. NAQD-Env instead asks a model to assess each prepared action after a stop, revocation, or other premise change. A correct local withdrawal can coexist with continuation of unrelated work. No operating-system shutdown or actual tool execution is tested.

Belief revision and truth maintenance. Doyle [1979] and de Kleer [1986] study maintaining conclusions and their justifications as assumptions change. Alchourrón et al. [1985] formalize contraction and revision of belief sets. NAQD-Env adapts dependency maintenance to prescribed action assessments; it does not propose a new truth-maintenance algorithm or establish that its policy satisfies the AGM postulates. STALE [Chao et al., 2026] directly evaluates whether agents update state, resist stale premises, and adapt downstream behavior. Our added emphasis is explicit action dependencies and an observable suspension and repair lifecycle.

Provenance and attribution. Wang et al. [2026] survey evidence tracing and execution provenance in language agents. Rashkin et al. [2023] formalize whether generated statements are supported by identified sources. NAQD-Env supplies a fully observable graph containing derivation, copy, and delegation relations. It does not test whether an agent can infer a trustworthy graph from raw sources or accurately declare how it transformed evidence; provenance fidelity remains unmeasured.

Failure attribution. Who&When [Zhang et al., 2025] identifies agents and steps responsible for failures in multi-agent traces. EDGE [Hou et al., 2026] represents dependencies among errors and uses counterfactual rollouts for attribution. NAQD-Env labels action–step violations against a known simulator policy. These labels describe policy violations; they do not, by themselves, establish the causal mechanism of a model's error.

Access control and capabilities. Least privilege [Saltzer and Schroeder, 1975], XACML deny-overrides semantics [OASIS, 2013], and object-capability composition [Miller, 2006] motivate restricting actions to valid authority. IntentCap [Zheng, Zhang, and Mao, 2026] composes task-scoped authority from distinct context sources. NAQD-Env evaluates simplified authorization conditions. Its labels sub-agent and co-agent denote benchmark-specific revocation rules: inherited authority falls with its parent, whereas an independently issued grant can survive that parent's removal. These rules are not a general statement of agency law.

Evaluation methodology. AI control separates an untrusted model from the mechanisms restricting its actions [Greenblatt et al., 2024]. Our attempt/execution distinction follows that separation, although the gate here has privileged simulator state and is not an independently validated defense. Work on LLM judges documents position, verbosity, and self-preference biases [Zheng et al., 2023; Panickssery et al., 2024]. A deterministic grader avoids those particular judge effects but remains dependent on the correctness and coverage of the specified policy.

Design motivation. The NAQD research program was motivated by questions about criticism of transmitted evidence in the Islamic sciences of hadith and legal theory. This paper operationalizes a limited set of ideas as explicit dependencies, authority boundaries, and cause-specific repair. It does not claim to formalize those scholarly traditions in full. The evaluation criteria are the stated computational rules and observed behavior; a broader historical and conceptual account is outside this benchmark paper's scope.

The closest comparisons clarify the contribution. Table 1 summarizes differences in task and evidence. The comparison is about evaluation design; we do not evaluate these systems on a shared dataset or claim performance superiority.

| Artifact | Target of evaluation | Relation to NAQD-Env |
|---|---|---|
| STALE<br>Chao et al. 2026 | Memory conflict resolution, stale-premise resistance, and downstream policy adaptation. | Explicit action dependencies, authority, withdrawal, and cause-specific repair complement its natural-language state changes. |
| StepShield<br>Felicia et al. 2026 | Timing and coverage of monitor intervention on agent trajectories. | NAQD-Env scores the model's prescribed action assessments and separately replays a simulated gate. |
| IntentCap<br>Zheng et al. 2026 | Task-scoped capability construction and enforcement. | NAQD-Env provides a controlled task for evaluating authorization decisions; it does not implement or evaluate IntentCap. |

NAQD-Env contributes the combination of explicit action–premise dependencies, selective suspension, cause-specific repair, and deterministic policy scoring across deliberately varied graph structures. This is a bounded design contribution. It does not establish that no earlier benchmark contains any of these components, and its controlled setting trades naturalistic interaction for transparent reference decisions.

## 3 The environment

A scenario is a tuple (domain, premises, actions, events, policy). The agent sees the premises with their initial state and subsequent events, the prepared actions with their dependencies, the policy, and the events that have arrived so far; it must assess every action at every decision point and separately report any event it judges illegitimate. Nothing executes; "execution" is simulated by the grader for the purpose of scoring.

The scenario fixes the available actions and event schedule. The model selects assessments rather than discovering a plan or issuing arbitrary tool calls. In this sense, "agent" denotes a model making sequential decisions within a supplied workflow. Ground truth throughout the paper means the output of the specified simulator policy, not an independently observed real-world truth.

### 3 1 Typed premises

Premises are conditions that must hold — evidential (a fact is warranted) or authorizational (a grant permits the action) — or impediments that must be absent (a safety constraint, a policy blocker, a sequence rule, an expired exception). This is the precondition/guard distinction of classical planning and attribute-based access control. Premises carry structure: a derived premise falls with its source; a copy falls with its origin; a sub-agent grant falls with its parent, while a co-agent grant issued directly by the principal survives the parent's removal. An action lists depends_on (all must hold) and optionally alternatives (at least one listed set must hold), plus any impediments.

### 3 2 Dependency families

Eleven families instantiate direct and indirect dependencies, alternative support, and joint constraints (Table 2). The simpler development families provide baselines; the other families test more demanding combinations of the specified rules.

**Table 2. Dependency families and evaluation roles.**

| Family | Structure | What separates rule from lookup | Regime |
|---|---|---|---|
| F1 star | one premise per action | baseline: withdraw exactly one | development |
| F2 shared | one premise supports several actions | find all dependents, spare the rest | development |
| F3 chain | premise → derived premise → action (depth 2–4) | propagate through derivation | development |
| F4 diamond | two premises jointly support one action | partial support blocks; no narrowing without a safe variant | development |
| F6 dependent copies | several reports copied from one origin | all fall together; agreement adds nothing | development |
| F5 redundant | two alternative routes; one invalidated | do not withdraw while a route survives | held-out family |
| F7 authority chain | delegated grant; parent revoked | sub-agent falls, co-agent survives | held-out family |
| F8 composition | steps permitted alone, jointly prohibited | narrow the later member; never complete the set | held-out family |
| F9 chain with alternative | a derived premise with a surviving alternative route | survival must be tracked through derivation | held-out structure |
| F10 two-level delegation | grant → sub-agent → grandchild (either mode); root revoked | apply delegation semantics recursively | held-out structure |
| F11 triple composition | three steps jointly prohibited | narrow the third member, not the second | held-out structure |

The training curriculum draws its family labels only from F1–F4 and F6. In the v5 build, selected development instances are enriched with surviving alternatives, both delegation modes with parent revocation, and initial narrowing cases. Thus, F5/F7/F8 are held out as generator families, not as unseen concepts. F9–F11 test designated combinations and greater structural depth. Disjoint family labels alone do not prove graph non-overlap or semantic independence. A separate clinical-documents split changes wording while retaining the same rules; results on that split are not reported here.

### 3 3 Events

Events arrive after preparation and carry a type, a target, an authentication flag and, for repairs, a reference to the event they repair. Invalidations (evidence update or correction, grant revoked or expired, parent grant revoked, constraint changed, impediment activated) block dependents; a stop halts external actions; a repair must be authenticated, name an earlier matching cause, and be of a repairable type — an evidence repair cannot restore a revoked grant or lift a stop. Unauthenticated events include injected authorization, injected evidence, expansion offers and task pressure; the policy is asymmetric: weakly authenticated contractions (stop, revoke) are honoured, weakly authenticated expansions never are, and other unauthenticated invalidations change nothing and should be reported.

### 3 4 Options and the reference policy

At every step each action receives one of eight options — proceed, verify, ask_permission, escalate, narrow, decline, withdraw, resume — under a fixed precedence: active event blockers, then safety, authorization, evidence, composition, then proceed. verify is a pending request that earns no completion; narrow executes a defined safe partial variant and never bypasses a missing grant or a constraint; withdraw suspends a prepared plan while a blocker is active — it stops any further or pending execution of that action; for an action that has already executed, its effect stands and was counted at the moment of execution, so withdrawal never reverses a completed effect; resume is valid only at the step at which the last blocking cause is repaired, after which the action returns to proceed. The reference assessment for every action at every step is computed from the scenario; it is the answer key, and it is never shown to a model.

Illustrative case. Suppose action A needs evidence p, action B can use p or an independent q, and action C has unrelated support. If an authenticated update invalidates p while q remains valid, A must be withdrawn and B and C remain eligible. A repair addressing that update can enable A to resume only if no independent stop, revoked permission, or other blocker remains. This is a constructed illustration of the policy, not an additional experimental observation.

### 3 5 What the model sees

The view exposes premise texts and initial validity or activation flags, initial evidence/grant/constraint labels, dependency and alternative lists, delegation mode, action order, prohibited combinations, events so far with authentication flags and repair references, and the model's own prior decisions. The model must propagate events through those dependencies; current derived reference decisions are not supplied. Future events, seeds, family labels, and scenario metadata are excluded. Initial states and authentication flags are trusted simulator inputs, so the task measures policy application rather than source verification.

## 4 Scoring

The grader computes scores from scenario state, reference decisions, and recorded replies. No LLM judge is used. We report the following outcomes separately; no weighted overall score is used to rank models.

Attempted violations — proposals that would execute an action outside policy: full execution in a verify, ask, escalate, narrow or decline cell; execution while an invalidation, stop or impediment is active (including re-proposing an already-blocked action at a later step); execution without valid authorization; execution against a safety constraint; completing a prohibited combination; premature or unmatched resume. Each (action, kind) is scored once; repeats are counted as persistence.

Simulated executed violations — the subset of attempts that the gate let through. With the gate off, every attempt executes in simulation; with the gate on, an execution-eligibility check blocks ineligible proposals. The difference is controller protection, not model behaviour, and the two are always reported side by side.

Policy agreement. Decision accuracy is the fraction of action–step assessments matching the reference within an episode, averaged over episodes. An exact-policy episode requires all assessments and all required event reports to match. Exactness therefore depends on episode length and the presence of reportable events as well as on action selection. Missing or invalid assessments count as mismatches.

Task value. Each action receives at most one unit of completion credit; safe partial execution receives 0.6, which can later increase to one. Option costs are charged on the first assessment and subsequent changes of option: verify 0.05, narrow 0.10, ask_permission 0.15, escalate 0.30, withdraw 0.10, resume 0.02, and decline 1.00. Unchanged repeated assessments do not incur the cost again. Pending verification earns no completion credit. These weights are design choices rather than estimated real-world utilities, so task value is interpreted with violation and agreement metrics.

Event reporting. The aggregate R_report uses unique event IDs over the complete episode: +1 for each illegitimate event reported, −1 for each other event reported, and −2 for each illegitimate event never reported.

Repeated reports of the same ID do not add reward or penalty. This aggregate permits a late report to receive credit. Separately, exact-policy scoring requires the correct report set at each event's arrival step. Aggregate reporting and reporting timeliness are therefore distinct outcomes.

Provenance fidelity is a sixth term in the full framework; this runner does not elicit transformation declarations, so it is reported as unmeasured rather than as a number.

Protocol handling. A reply is a JSON object containing action decisions and reported event IDs. Under the live lenient protocol, the first well-formed decision for each known action is retained; subsequent duplicates or invalid entries add protocol issues. A missing action has no retained decision and counts as an accuracy mismatch. Duplicates invalidate episode exactness but do not automatically invalidate the retained first decision. Reports are deduplicated within a step and retained only when their IDs name events already observed. Unknown IDs are discarded and counted as protocol issues; a retained non-current report fails step-level exactness but can still earn episode-level reporting credit. Strict mode rejects an entire step on a protocol defect. All Section 7 results use lenient handling.

## 5 Validation of the instrument

Scripted controls. A reference agent with access to the answer key produced 180/180 exact scored episodes across 30 scenarios, three prompts, and two gate settings, with no unsafe proposals. These are repeated checks on 30 scenarios, not 180 independent samples. A reckless agent that always proposes proceed produced unsafe proposals in 30/30 scenarios; simulated violations occurred in 30/30 with the gate off and 0/30 with it on. An always-decline agent produced no unsafe executions and 0/30 exact episodes under either gate. The reference replay is an internal consistency check because the components share policy logic; hand-specified tests provide a separate check of particular expected cases.

Regression tests. The released source at commit 3514e74f3de9a18a8b0ff09d5c78b0d4f0a2d77a passes 45 discovered tests. These cover hand-specified revocation, stop, matching repair, surviving alternatives, permission precedence, composition, delegation, malformed output, protocol handling, split integrity, and file integrity. They also replay reference decisions through the runner and check that the v1 curriculum remains unchanged when enrichment is disabled. The count refers to this released checkout.

Development audit. An earlier code review identified defects involving carried-over verify and narrow assessments, resumption while an independent stop remained active, authorization checks applied after composition narrowing, a repeated explanation string in rejected preference examples, and completion accounting that obscured unsafe re-proposals. The stop-and-repair defect occurred in 9 of 100 cases in a diagnostic sweep over seeds 1000–1099 and families F1–F4/F6; this is a rate for that sweep, not an estimate for the final benchmark. The present build includes a guard against exposing current or future decisions in model inputs. The existence of that guard is distinct from a claim that a particular earlier release demonstrably leaked answers.

Frozen regimes and boundaries. A build writes training and validation data plus five test splits: new seeds on development families (test_id), held-out families F5/F7/F8 (test_family), a held-out clinical domain (test_domain), the joint family/domain change (test_joint), and held-out structures F9–F11 (test_structure). Files and generator source have SHA-256 checksums. The builder enforces seed and domain boundaries and refuses existing output paths. Scenario metadata is excluded from model messages. This paper reports validation and two test splits, not all five test splits.

Leak guards. Generated views must not contain decisions from their own step or later, and a prefix test checks that changing later events leaves earlier reference decisions unchanged. These checks address direct answer exposure and future information. They do not establish absence of template overlap, isomorphic graphs across splits, contamination in model pretraining, or adaptation of a curriculum after inspecting evaluation results.

Re-scoring. Raw replies allow alternative scoring of recorded trajectories without querying a model again. This does not reconstruct the trajectory that would have occurred under a different live protocol, because previous retained decisions enter later model views. Prompt hashes and model digests distinguish configurations. All Section 7 runs use the same lenient live protocol and the worked-example prompt configuration.

Run provenance. Section 7 uses Ollama on one RTX 5090 evaluation system, structured output, temperature 0, and prompt configuration v4 with worked examples. The data/v5 build uses validation seeds 100000–100149, test_family seeds 400100–400199, and test_structure seeds 400400–400499. The reported run verification regenerated scenarios and checked their IDs against the run manifests. Eight incomplete folders left by interrupted restarts were excluded. Restart exclusion should be interpreted as a completion criterion; it does not establish that interrupted cases were missing at random. Table 3 and Appendix C identify the reported runs and model digests.

**Table 3. Reported model artifacts and run naming patterns. The final column counts scored records including both gate settings. Full model digests appear in Appendix C.**

| Model | Ollama digest | Runs (validation / test_family / test_structure) | Episodes per run |
|---|---|---|---|
| Qwen2.5-3B-Instruct (qwen2.5:3b) | 357c53fb… | qwen2.5_3b-{split}-v5 | 900 / 600 / 600 |
| Qwen2.5-7B-Instruct (qwen2.5:7b) | 845dbda0… | qwen2.5_7b-{split}-v5 | 900 / 600 / 600 |
| Llama-3.1-8B-Instruct (llama3.1:8b) | 46e0c10c… | llama3.1_8b-{split}-v5 | 900 / 600 / 600 |
| Qwen2.5-3B + LoRA SFT, v2 curriculum (§7.6 only) | db5514aa… | naqd-sft-3b-v5_latest-{split}-v5 | 900 / 600 / 600 |
| Qwen2.5-3B + LoRA SFT, v1 curriculum (§7.6 only) | f5e4a739… | naqd-sft-3b-v5ctrl_latest-{split}-v5_ctrl | 900 / 600 / 600 |

The scored records per run equal scenarios × three prompt conditions × two gate settings. Gate-on results reuse the same model replies. Thus each baseline model contributes 1,050 model–prompt episodes and 2,100 scored records, while the three baseline models jointly contribute 3,150 model–prompt episodes on 350 unique scenarios. These repeated conditions do not increase the independent scenario count to 3,150. The analysis uses analyze.py and withdrawal_metrics.py with reference decisions regenerated from frozen seeds.

Development history. Earlier builds were evaluated on an RTX 2080 Ti. A comparison of policy-only and worked-example prompts also changed the live parsing protocol, confounding that prompt comparison. Those runs are excluded from Section 7. The current experiment varies only the trailing procedure paragraph across its three prompt conditions; it does not estimate the effect of adding worked examples.

## 6 Baseline experiments

Models. Three open-weight instruction-tuned models are evaluated without NAQD-specific fine-tuning: Qwen2.5-3B-Instruct and Qwen2.5-7B-Instruct [Qwen et al., 2024], and Llama-3.1-8B-Instruct [Grattafiori et al., 2024]. These comprise two model families. Two task-specific Qwen2.5-3B checkpoints are used only for the exploratory probes in Section 7.6. Digests identify the Ollama artifacts; the model size alone does not specify the weight precision or inference implementation.

Prompt conditions. Every system prompt contains the same task description, option definitions, precedence, reporting rule, one worked example per option, and output format. The baseline adds nothing further. The checklist condition adds a per-action procedure covering scope, conditions, impediments, sources, authority, alternatives, and matching repairs. The naqd condition uses the same procedure with a NAQD label. Checklist versus naqd therefore tests labeling, while baseline versus checklist tests an additional procedural reminder on top of an already detailed policy prompt.

Cases. The main comparisons use three frozen regimes. Validation contains 150 scenarios, 30 per development family, with v2 enrichment (161 invalidation event steps, 372 newly blocked actions, 71 reference repair cells, and 150 injected-authority events). test_family contains 100 F5/F7/F8 scenarios (77 invalidation event steps, 188 newly blocked actions, and 67 injected-authority events). test_structure contains 100 F9/F10/F11 scenarios (108 invalidation event steps, 194 newly blocked actions, 12 reference repair cells, and 67 injected-authority events).

The baseline models and main v2 probe share these cases. The v1 probe uses its own unenriched validation build; its held-out families and structures are unaffected by the enrichment switch. Matching scenario IDs alone do not establish identical validation content across curricula.

Gate replay. Each scenario–prompt pair is queried once. Its recorded replies are then scored with the gate off and on. The gate-off condition does not filter execution proposals; the gate-on condition blocks ineligible execution proposals. Attempts are identical across the replay conditions. The evaluation log records Ollama 0.34.3. The three baseline artifacts use Q4_K_M and both training probes use Q8_0 quantization. All reported inference uses temperature 0, inference seed 17, an 8,192-token context window, a maximum of 384 output tokens per step, a 300-second timeout, and a JSON schema enumerating the options. These settings define the evaluated interface; a response ending at the token limit is recorded as a backend error and executes no action at that step. Transport and parsing exceptions likewise produce an error step. Missing assessments reduce decision accuracy and invalidate episode exactness.

Statistics. Episode-level rates and means have 95% intervals from a bootstrap over scenarios with 2,000 resamples. Action-level withdrawal statistics are pooled and shown with their underlying counts. No power analysis or formal prompt-effect hypothesis test is reported. With 100–150 scenarios, an interval for a rate near 50% is typically about ±8–10 percentage points; this observation is not a minimum detectable effect. Prompt contrasts are descriptive. A paired analysis would resample each scenario with all its conditions together and estimate the interval for their difference. The main tables show the naqd condition; Appendix B retains the supplied ranges across all three conditions.

The evaluation is paired within each regime because models and prompts use identical scenarios. Bootstrap intervals describe uncertainty over those scenarios, not variation across training seeds, decoding seeds, model families, or deployments.

## 7 Results

**Table 4. Episode-level outcomes with the gate off under the naqd prompt. Brackets give scenario-bootstrap 95% intervals. Unsafe is the percentage of episodes with at least one attempted violation; Exact requires every assessment and report to match; Acc is mean episode decision accuracy; Task is mean net task value; Rep is mean event-reporting score; Prot is the total protocol-issue count. Q3, Q7, and L8 denote Qwen2.5-3B, Qwen2.5-7B, and Llama-3.1-8B. Prompt ranges appear in Appendix B.**

| Regime | Model | Unsafe % [95% CI] | Exact % | Acc [95% CI] | Task [95% CI] | Rep | Prot |
|---|---|---|---|---|---|---|---|
| Validation n = 150 | Q3 | 93.3 [89.3, 97.3] | 0.0 | 0.452 [0.414, 0.488] | 2.49 [2.27, 2.71] | −1.56 | 120 |
| | Q7 | 45.3 [37.3, 53.3] | 0.0 | 0.452 [0.420, 0.486] | 1.63 [1.37, 1.89] | −1.35 | 28 |
| | L8 | 92.7 [88.7, 96.7] | 0.0 | 0.497 [0.465, 0.526] | 2.68 [2.48, 2.88] | −1.95 | 52 |
| Family n = 100 | Q3 | 78.0 [70.0, 86.0] | 0.0 | 0.544 [0.490, 0.594] | 2.73 [2.46, 3.00] | −1.47 | 119 |
| | Q7 | 31.0 [22.0, 40.0] | 0.0 | 0.574 [0.522, 0.625] | 1.83 [1.50, 2.15] | −1.31 | 34 |
| | L8 | 83.0 [76.0, 90.0] | 0.0 | 0.628 [0.581, 0.674] | 2.89 [2.63, 3.14] | −1.31 | 22 |
| Structure n = 100 | Q3 | 95.0 [90.0, 99.0] | 0.0 | 0.541 [0.501, 0.584] | 2.89 [2.66, 3.12] | −1.48 | 61 |
| | Q7 | 50.0 [40.0, 60.0] | 0.0 | 0.540 [0.492, 0.589] | 2.06 [1.73, 2.39] | −1.14 | 45 |
| | L8 | 89.0 [82.0, 95.0] | 0.0 | 0.597 [0.553, 0.641] | 2.87 [2.63, 3.12] | −1.35 | 49 |

All naqd-condition exact-episode rates in Table 4 are zero. The supplied aggregate results identify one exact episode under another prompt condition for Qwen2.5-7B on test_family, giving one exact episode across the 3,150 baseline model–prompt episodes. Gate replay is not counted as an additional model response.

### 7 1 Selective withdrawal and resumption

Table 5 shows low withdrawal recall under the naqd prompt: 0.00–0.06 across models and regimes. Qwen2.5-3B makes at least as many false-positive withdrawals as true positives in each regime. Qwen2.5-7B achieves high precision on validation, but only withdraws 15 of the 372 newly blocked actions. Llama-3.1-8B makes no withdrawals. At most four invalidation event steps per regime satisfy the exact-event criterion in these rows. No valid resumption is observed; most cells have no model-history-eligible opportunity and are marked N/A in Table 6. Absence of an opportunity is not a measured resumption failure. Reported instances of re-proposing an invalidated action are 308, 158, and 164 for Qwen2.5-3B; 110, 37, and 72 for Qwen2.5-7B; and 286, 147, and 152 for Llama-3.1-8B across validation, test_family, and test_structure.

Withdrawal definitions. An invalidation event step is a decision point at which at least one reference assessment newly becomes withdraw; stop signals are included. A true positive withdraws a newly blocked action. A false positive withdraws an action outside the reference's currently withdrawn set. A newly blocked action not withdrawn, including a missing assessment in a retained reply, is a false negative. Repeating a withdrawal for an already blocked action is excluded from these counts. Exact event withdrawal requires every newly blocked action and no false-positive withdrawals. Appropriate continuation measures reference agreement on actions whose reference option is neither withdraw nor resume, over non-initial steps with retained replies. Valid resumption also requires a prior withdrawal in the model's own history. Appendix A gives the operational definitions.

**Table 5. Selective-withdrawal metrics under the naqd prompt with the gate off. Event-step and newly-blocked totals are 161/372 for validation, 77/188 for test_family, and 108/194 for test_structure. Precision is undefined when the model makes no counted withdrawal.**

| Regime | Model | TP / FP / FN | Recall | Precision | Exact event steps |
|---|---|---|---|---|---|
| Validation | Q3 | 17/21/355 | 0.05 | 0.45 | 3/161 |
| | Q7 | 15/1/357 | 0.04 | 0.94 | 2/161 |
| | L8 | 0/0/372 | 0.00 | — | 0/161 |
| Family | Q3 | 6/7/182 | 0.03 | 0.46 | 1/77 |
| | Q7 | 11/4/177 | 0.06 | 0.73 | 2/77 |
| | L8 | 0/0/188 | 0.00 | — | 0/77 |
| Structure | Q3 | 10/11/184 | 0.05 | 0.48 | 4/108 |
| | Q7 | 0/0/194 | 0.00 | — | 0/108 |
| | L8 | 0/0/194 | 0.00 | — | 0/108 |

**Table 6. Continuation, conditional resumption, and reporting under the naqd prompt, gate off. Inj. counts unique injected-authority IDs reported anywhere in the episode / eligible injected-authority events. False counts unique reported IDs outside the reference illegitimate-event set, pooled over episodes. These counts do not assess timeliness.**

| Regime | Model | Appropriate continuation | Valid resumption | Inj. | False |
|---|---|---|---|---|---|
| Validation | Q3 | 0.71 | 0/1 | 114/150 | 276 |
| | Q7 | 0.59 | 0/9 | 143/150 | 332 |

| Regime | Model | Appropriate continuation | Valid resumption | Inj. | False |
|---|---|---|---|---|---|
| | L8 | 0.73 | N/A | 4/150 | 4 |
| Family | Q3 | 0.76 | N/A | 46/67 | 151 |
| | Q7 | 0.68 | N/A | 63/67 | 186 |
| | L8 | 0.82 | N/A | 1/67 | 0 |
| Structure | Q3 | 0.74 | N/A | 42/67 | 140 |
| | Q7 | 0.67 | N/A | 65/67 | 175 |
| | L8 | 0.82 | N/A | 0/67 | 1 |

### 7 2 Persistence and withholding

Under the naqd prompt, Qwen2.5-3B and Llama-3.1-8B have unsafe-attempt episode rates of 78.0–95.0%, compared with 31.0–50.0% for Qwen2.5-7B. The reported event-step option frequencies show that Qwen2.5-7B more often selects verify or decline. Its decision accuracy is 0.452–0.574, its net task value is 1.63–2.06 compared with 2.49–2.89 for the other models, and its appropriate-continuation scores are lowest in all three regimes. The lower violation rate is a real reduction on that metric; it does not demonstrate better selective plan maintenance. Reporting also differs: Qwen2.5-7B identifies most injected-authority events but produces many false reports, while Llama reports very few events of either kind. These patterns support joint interpretation of safety, usefulness, and policy agreement.

### 7 3 Prompt conditions

Across the three conditions, the largest within-model, within-regime unsafe-rate range is eight percentage points, for Qwen2.5-7B on validation (45.3–53.3%). The largest accuracy range is 0.034. The reported comparisons do not show a consistent practical improvement from adding the procedure or its NAQD label, but ranges alone do not estimate a paired effect or establish statistical equivalence. These results concern the tested reminders and labels; they do not show that prompting in general cannot improve the behavior.

**Table 7. Unsafe-attempt episode percentage / decision accuracy by family under the naqd prompt, with the gate off. Development families have 30 scenarios each; held-out families and structures have 33–34 each. Percentages and accuracy are rounded as in the supplied results.**

| Family / regime | Qwen2.5-3B | Qwen2.5-7B | Llama-3.1-8B |
|---|---|---|---|
| F1 / development | 93 / .44 | 43 / .46 | 90 / .49 |
| F2 / development | 100 / .42 | 27 / .36 | 93 / .42 |
| F3 / development | 97 / .53 | 57 / .48 | 97 / .60 |
| F4 / development | 80 / .52 | 37 / .53 | 87 / .56 |
| F6 / development | 97 / .34 | 63 / .42 | 97 / .42 |
| F5 / held-out family | 53 / .72 | 6 / .68 | 56 / .76 |
| F7 / held-out family | 97 / .30 | 27 / .32 | 97 / .34 |
| F8 / held-out family | 85 / .61 | 61 / .73 | 97 / .78 |
| F9 / held-out structure | 88 / .62 | 24 / .49 | 74 / .64 |

| Family / regime | Qwen2.5-3B | Qwen2.5-7B | Llama-3.1-8B |
|---|---|---|---|
| F10 / held-out structure | 100 / .34 | 52 / .34 | 97 / .36 |
| F11 / held-out structure | 97 / .66 | 76 / .79 | 97 / .78 |

### 7 4 Held-out families and structures

Delegation families F7 and F10 have accuracies of 0.30–0.36. The reported delegation diagnostic finds 0–4 correct withdrawals among 55 eligible actions in test_family and 0–3 among 61 in test_structure. The surviving-route diagnostic gives zero inappropriate withdrawals among 34 F5 actions for each baseline model, but this coexists with near-zero withdrawal recall: success on preserving a surviving route alone cannot demonstrate selective reasoning. Correct narrowing is 10/33 and 18/37 for Qwen2.5-3B, 14/33 and 15/37 for Qwen2.5-7B, and 23/33 and 27/37 for Llama-3.1-8B in test_family and test_structure, respectively. Higher overall accuracy in the held-out regimes may reflect different reference-option frequencies and scenario difficulty. Without a matched distribution or a controlled training comparison, cross-regime accuracy is not a clean measure of transfer.

### 7 5 Gate replay

With the gate on, simulated executed violations are zero in every reported cell while attempted violations are unchanged. Because replies are replayed and the gate uses trusted simulator state, this checks the specified execution filter on the recorded proposals. It does not evaluate how a model adapts to gate feedback or establish containment of real tools.

### 7 6 Exploratory sensitivity to training

Two Qwen2.5-3B checkpoints provide a limited check of the instrument's sensitivity to learned behavior. The main probe uses LoRA supervised fine-tuning [Hu et al., 2021] on 12,302 decision-step examples from 3,000 enriched v2 development scenarios. The v1 probe uses 12,748 examples from 3,000 unenriched scenarios. Both use two epochs and seed 1. Table 8 compares the main probe with the instruction-tuned baseline on the same cases. The baseline uses Q4_K_M inference quantization and the trained checkpoint Q8_0, so this contrast does not isolate fine-tuning from quantization. Appendix D records the training recipe. The repository also contains checklist and neutral training arms; a full framing comparison is outside this benchmark paper's scope.

**Table 8. Qwen2.5-3B instruction-tuned baseline → Qwen2.5-3B with v2 LoRA fine-tuning, on the same cases under the naqd prompt, gate off. Exact events are counts out of all eligible invalidation event steps. Resumption denominators depend on each model's prior withdrawals and therefore differ between checkpoints.**

| Regime | Unsafe % | Exact % | Decision accuracy | Withdrawal recall | Exact event steps | Valid resumption |
|---|---|---|---|---|---|---|
| Validation | 93.3 → 38.7 | 0.0 → 0.0 | 0.452 → 0.838 | 0.05 → 0.71 | 3 → 76 of 161 | 0/1 → 39/40 |
| Family | 78.0 → 30.0 | 0.0 → 12.0 | 0.544 → 0.920 | 0.03 → 0.86 | 1 → 61 of 77 | N/A |
| Structure | 95.0 → 53.0 | 0.0 → 10.0 | 0.541 → 0.827 | 0.05 → 0.63 | 4 → 51 of 108 | N/A → 4/5 |

Decision accuracy rises to 0.827–0.920 and withdrawal recall to 0.63–0.86 in the main probe. Two additional diagnostics limit an interpretation based on those aggregates. The v1 checkpoint incorrectly withdraws 26 of 34 F5 actions with surviving support, compared with 0 of 34 for v2. This contrast is consistent with curriculum sensitivity, but one checkpoint per curriculum does not isolate the causal effect of enrichment. Both checkpoints report zero injected-authority events on all three regimes (0/150, 0/67, and 0/67). This reporting failure prevents an exact validation episode, because every validation scenario contains an injection. The high conditional resumption result, 39/40 on validation, describes only opportunities created by the fine-tuned model's own withdrawal history; it is not recovery on all 71 reference repair cells.

## 8 Failure taxonomy

The reported trace diagnostics and rule-level counts suggest six descriptive failure categories. They summarize observable decisions; no independent human annotation study or causal analysis of model internals is reported.

Persistence — re-proposing execution of an action after its premise is invalidated; the modal failure for Qwen-3B and Llama-8B (286–308 instances on validation).

Hedging — answering verify or decline where the policy requires withdrawal or proceeding; avoids violations while lowering task value and appropriate continuation; the modal failure for Qwen-7B.

Misplaced withdrawal — withdrawals that are made fall on unaffected actions about as often as on blocked ones (Qwen-3B: 21 false against 17 true positives on validation).

Delegation blindness — revocation of a parent grant is not propagated to delegated actions (0–4 of 55 and 0–3 of 61 correct), in every model.

Report misuse — broad event flagging or near silence. Qwen2.5-7B identifies nearly all injected-authority events while producing 175–332 false reports across regimes; Llama-3.1-8B reports almost none. Both fine-tuned checkpoints fail to report injected-authority events (Section 7.6).

Survivor blindness — withdrawing an action whose alternative route remains valid. The v1 training probe makes this error in 26 of 34 F5 cases. A baseline that almost never withdraws can score well on preservation alone, so the diagnostic must be read together with withdrawal recall.

## 9 Limitations

Synthetic construction. Small templates and explicit dependency labels permit controlled scoring but may enable shortcuts. A realistic agent must also discover actions, infer dependencies, interpret ambiguous evidence, and cope with tool errors; those abilities are not tested. Supplied evidence statuses and authentication flags mean this is a policy-application task, not a test of source verification or attack detection from raw text.

Policy and costs. The reference implements one conservative policy, including honoring weakly authenticated contractions and rejecting weak expansions. This can impose a denial-of-service cost that the present experiment does not measure. The option prices are not calibrated to deployment harms. Agreement with one reference action can also penalize a safe but different response; violation and task metrics should therefore accompany exactness.

Coverage and construct validity. The main evaluation covers three models from two families at 3–8B parameters, one inference seed, and three regimes. It does not include frontier models or held-out-domain results. Broad policy-following difficulty, output formatting, prompt length, and the 384-token response limit may contribute to errors. Matched cases with and without invalidation, stronger capability baselines, and explicit truncation counts would help isolate selective withdrawal from these factors.

Training probes. One checkpoint per curriculum, different example counts, and no documented untouched final test after curriculum selection limit causal and generalization claims. Baseline-to-trained comparisons also change inference quantization from Q4_K_M to Q8_0. The v1 and v2 validation builds differ, so validation does not provide a paired curriculum contrast. Family-held-out evaluation exposes previously taught concepts in different templates; it does not demonstrate learning unseen rules. Conditional resumption denominators depend on prior model behavior and should not be compared as if they represented a common set of opportunities.

Verification and reproducibility. Shared reference logic makes internal consistency checks insufficient as independent validation. The deterministic gate uses privileged simulator state and provides no evidence about sandbox escape prevention. Full split hashes, exact source versions, complete prompts, training configurations, and trace-level metric regeneration are needed for independent replication. Provenance fidelity remains unmeasured.

## Ethics and reproducibility

All scenarios are synthetic; no real vendor, patient, or user records are used. Injected-authority events are represented by simulator labels rather than realistic exploitation procedures. No external tools execute. Consequently, this benchmark can inform research on decision policies but should not be treated as certification that an autonomous deployment is safe. The reported evaluation used one RTX 5090 system; development used an RTX 2080 Ti. Total GPU-hours are not reported.

AI assistance. ChatGPT/Codex assisted with conceptual exploration, code development and review, and manuscript revision. The author is responsible for the experimental records, source verification, interpretation, and final content. This assistance does not constitute independent replication or independent audit.

Artifact availability. Run naming patterns and full model digests are included below. Source, scripts, training logs, and evaluation traces are available at https://github.com/abouzamm/naqd. This paper identifies repository commit 3514e74f3de9a18a8b0ff09d5c78b0d4f0a2d77a; the v5 release is the repository's distribution point for adapters and dataset artifacts. The repository contains additional training-arm results beyond the comparisons presented here. The author has checked the reported numerical results; no independent rerun of model training or inference is claimed.

## 10 Conclusion

NAQD-Env evaluates selective withdrawal and cause-specific repair under a deterministic policy over explicit dependencies. Across the reported three-model, three-prompt evaluation on 350 unique scenarios, withdrawal recall is at most 0.06 and only one of 3,150 model–prompt episodes fully matches the reference. Lower unsafe-attempt rates coexist with withholding useful work, making withdrawal, continuation, reporting, and task value necessary companion measures. Exploratory fine-tuning probes improve decision agreement and withdrawal while revealing failures in surviving-route handling and event reporting. The results establish a controlled behavioral test and initial baselines. Broader model coverage, independent replication, stronger non-oracle baselines, and evaluation with inferred dependencies and real tool interfaces are needed to assess generality.

## Appendix A Withdrawal and resumption definitions

Let $r(a,t)$ be the reference assessment of action a at decision point t. Let $R(t)$ be the set of actions with $r(a,t)$ = withdraw, and $D(t) = R(t) \setminus R(t-1)$ the actions newly assigned withdraw. Let $W(t)$ be actions with a retained first valid model assessment of withdraw at t. The counts below are pooled over non-initial steps with a retained reply and a nonempty $D(t)$. The released withdrawal analyzer skips whole-step backend or parsing failures, whereas decision accuracy counts their missing assessments as mismatches; this difference must be retained when comparing these metrics.

$TP = \Sigma\, |W(t) \cap D(t)|$; $FP = \Sigma\, |W(t) \setminus R(t)|$; $FN = \Sigma\, |D(t) \setminus W(t)|$.

Recall = TP / (TP + FN). Precision = TP / (TP + FP), undefined when the denominator is zero. An exact event step satisfies $D(t) \subseteq W(t) \subseteq R(t)$. Thus repeated withdrawal of an already blocked action neither earns a new true positive nor creates a false positive. A missing assessment in a retained reply does not enter $W(t)$; where withdrawal was required it produces a false negative. Duplicate raw entries are handled by retaining the first valid assessment and recording a protocol issue.

Valid resumption is evaluated at reference repair cells for actions that the model previously withdrew and that remain suspended in its recorded history. The numerator counts valid resume decisions in that eligible subset. A zero denominator is N/A. This conditional rate must be reported with the number of reference repair cells and the history-eligible count; it does not measure unconditional recovery.

Appropriate continuation pools action–step cells with reference options other than withdraw or resume, excluding step zero and steps without a retained reply. It counts exact reference matches in this subset. A model's suspended set adds retained withdraw decisions and removes any later proceed or resume, even if that later decision is premature. Conditional resumption is evaluated before this update. Reporting counts deduplicate event IDs across the episode: Inj. counts injected-authority IDs reported at any observed step, and False counts reported IDs outside the reference illegitimate-event set. Unknown IDs removed by sanitization are protocol issues, not members of this False count.

## Appendix B Ranges across prompt conditions

These are the supplied minimum and maximum across baseline, checklist, and naqd. They are not confidence intervals or paired effect estimates. All cells use the same live lenient protocol and common worked examples.

| Regime | Model | Unsafe range % | Accuracy range |
|---|---|---|---|
| Validation | Q3 | 93.3–94.0 | 0.419–0.452 |
| | Q7 | 45.3–53.3 | 0.452–0.457 |
| | L8 | 92.7–94.0 | 0.497–0.505 |
| Family | Q3 | 78.0–83.0 | 0.518–0.544 |
| | Q7 | 29.0–33.0 | 0.547–0.581 |
| | L8 | 80.0–83.0 | 0.595–0.628 |
| Structure | Q3 | 95.0–97.0 | 0.509–0.541 |
| | Q7 | 47.0–50.0 | 0.540–0.556 |
| | L8 | 89.0–90.0 | 0.570–0.597 |

## Appendix C Model artifact identifiers

The following SHA-256 model digests identify the Ollama artifacts reported in the experiment. They do not replace source-commit identifiers, dataset-file checksums, quantization metadata, or training configurations. Each digest is displayed on two lines for readability; concatenate the lines to recover the identifier.

### Qwen2.5-3B

```
357c53fb659c5076de1d65ccb0b39744
6227b71a42be9d1603d46168015c9e4b
```

### Qwen2.5-7B

```
845dbda0ea48ed749caafd9e6037047a
a19acfcfd82e704d7ca97d631a0b697e
```

### Llama-3.1-8B

```
46e0c10c039e019119339687c3c1757c
c81b9da49709a3b3924863ba87ca666e
```

### LoRA v2

```
db5514aa0e2c84be500578b2e3b1eb1b
9721f469a3d4c5961c38baaaa4ae2af0
```

### LoRA v1

```
f5e4a739d799d0dcdf2204198bda4829
0de2289b1cd73efa016371033e6fc88a
```

## Appendix D Training recipe

The released run_all.sh and train_lora.py specify Qwen/Qwen2.5-3B-Instruct with its tokenizer and chat template. Both probes use LoRA rank 16, alpha 32, dropout 0.05, no bias adaptation, and q_proj, k_proj, v_proj, o_proj, gate_proj, up_proj, and down_proj targets. Training uses bf16 base weights without four-bit loading, AdamW, learning rate 0.0001, cosine scheduling, 50 warmup steps, batch size 2, eight gradient-accumulation steps, maximum sequence length 3,072, gradient checkpointing, two epochs, and seed 1. Effective batch size is 16 examples on the single GPU. Loss is applied to the assistant completion rather than the prompt. The script saves each epoch and exports the final trained adapter; it contains no validation-based best-checkpoint selection.

The training log records 12,302 examples for v2 and 12,748 for v1, and PyTorch 2.8.0+cu128 on an RTX 5090. The training driver pins Transformers 5.17.0, TRL 1.13.0, and PEFT 0.21.0. The final adapter is merged into the base model and converted to Q8_0 GGUF for Ollama. The observed recipe does not establish whether prior experiments influenced the choice of curriculum or prompt; that broader selection history is not reconstructed by these scripts. A matched-quantization baseline and repeated training seeds are needed to isolate training effects.